\documentclass{article}
\usepackage{ijcai22}

\usepackage{times}
\usepackage{soul}
\usepackage{url}
\usepackage[hidelinks]{hyperref}
\usepackage[utf8]{inputenc}
\usepackage[small]{caption}
\usepackage{graphicx}
\usepackage{amsmath}
\usepackage{amsthm}
\usepackage{booktabs}
\usepackage{algorithm}
\usepackage{algorithmic}
\usepackage{multirow}

\usepackage{subcaption}
\usepackage{soul}
\usepackage{color}
\usepackage{amsfonts}
\usepackage{multirow}
\usepackage{siunitx}
\title{S-Omninet: Structured Data Enhanced Universal Multimodal Learning Architecture}

\author{
Ye Xue$^1$
\and
Diego Klabjan$^2$\And
Jean Utke$^{3}$%\and
\affiliations
$^{1,2}$Northwestern Universion\\
$^3$Allstate Insurance Company\\
\emails
ye.xue@u.northwestern.edu,
d-klabjan@northwestern.edu,
jutke@allstate.com %,Keith.Matera,dzorn,kinde\}@allstate.com
}
\begin{document}

\maketitle

\begin{abstract}
Multimodal multitask learning has attracted an increasing interest in recent years. Singlemodal models have been advancing rapidly and have achieved astonishing results on various tasks across multiple domains. Multimodal learning offers opportunities for further improvements by integrating data from multiple modalities. Many methods are proposed to learn on a specific type of multimodal data, such as vision and language data. A few of them are designed to handle several modalities and tasks at a time. In this work, we extend and improve Omninet, an architecture that is capable of handling multiple modalities and tasks at a time, by introducing cross-cache attention, integrating patch embeddings for vision inputs, and supporting structured data. The proposed Structured-data-enhanced Omninet (S-Omninet) is a universal model that is capable of learning from structured data of various dimensions effectively with unstructured data through cross-cache attention, which enables interactions among spatial, temporal, and structured features. We also enhance spatial representations in a spatial cache with patch embeddings. We evaluate the proposed model on several multimodal datasets and demonstrate a significant improvement over the baseline, Omninet. 
\end{abstract}

\section{Introduction}\label{intro}

% 1. Trend of multimodal learning, increasing use cases
% 2. Motivation of having one model (multimodal, easy to accommodate new data type, less human intervention) 
% 3. One model to learn them all, Omninet
% 4. What need to be improved in Omninet (not support for structured data, especially structured data of various length; one direction communication between caches)
% 5. Our method
% 6. summarize contribution

Deep learning yielded great success in the last decade on tasks across many domains with various types of unstructured data such as images and text. Specific models are carefully designed and tailored for a particular type of data. For example, image recognition models \cite{he2016deep} are built with convolutional neural networks (CNNs), while recurrent neural network (RNN) has shown success for sequential data. Although some recent works use CNN in language tasks and transformer on image tasks \cite{dosovitskiy2020image}, these tasks are still single-modal tasks. 

In recent years, multimodal learning attracted an increasing interest, as the single modal models are advancing rapidly and achieving astonishing results on various tasks across multiple domains, such as vision \cite{dosovitskiy2020image} and language \cite{devlin2018bert}. Multimodal tasks are more common in complex tasks and multimodal learning has been shown more effective than models that used only single modalities in several fields, such as healthcare, where models are trained to utilize medical images and electronic health records \cite{huang2020fusion,li2021multimodal}, and autonomous driving where intelligent systems are built to process various signals in different modalities \cite{feng2020deep,prakash2021multi}.

However, existing multimodal learning models are still usually tailored to specific tasks or modalities and are not easily extended to accommodate new types of data. Different models for different tasks need to be separately designed and maintained. In business, new tasks may be added over time as a business grows and new types of data may also need to be considered to best utilize the extra information. One architecture that handles multiple modalities and multiple tasks becomes increasingly appealing.

MultiModel \cite{kaiser2017one} is probably the first attempt at building a single model that can solve tasks across multiple domains. It consists of several modality nets, an encoder, and a decoder. Each modality net handles one type of input data. A mixer module gathers encodings from modality nets and feeds them to the decoder. However, for a single task, MultiModel does not have support for inputs having more than one modality, such as Visual Question Answering (VQA). In order to address this challenge, another multi-model multi-task architecture, Omninet, is proposed \cite{pramanik2019omninet}. Similar to modality-nets, Omninet has a visual peripheral to encode images and videos and several language peripherals to encode text data in different languages. Inputs in different modalities are further encoded into temporal and spatial caches, which are fed into a transformer-based decoder. 

Omninet is shown to have competitive performance in several tasks compared with state-of-the-art models. However, there are still a few shortcomings of Omninet. First, each modality in Omninet is encoded in a separate stream. Existing works \cite{tsai2019multimodal,tan2019lxmert,lu2019vilbert} have shown that it is beneficial for multimodal learning models to encode one modality with the information of other modalities. Second,  it lacks support for structured data. In many real world applications structured data play an important role, even in a multimodal scenario. For example, similar medical image findings may suggest different diagnoses given different laboratory test results \cite{huang2020fusion}. Unfortunately, many of the practical applications are built on proprietary data sets making this specific topic less accessible for academic research. However, there are countless cases where full context, in the form of structured and unstructured, is critical for making accurate decisions. A straightforward way of extending Omninet to utilize structured data is concatenating structured features with the final encoding vector of the other modalities. However, such a late fusion mechanism ignores the potential informative interactions between structured and unstructured data. Furthermore, it would not deal with a varying number of structured data sources, because naive concatenation fixes the number of structured data sources in the network setup. 

In this work, we extend Omninet with a design of a structured peripheral and structured cache. Instead of common encoding methods that encode categorical structured features into one vector, such as one-hot encoding, we encode structured data using entity embeddings \cite{guo2016entity}. We store encodings of structured data into the structured cache. It interacts with other caches through a cross cache attention mechanism, which we propose to enhance the encodings by considering those from other caches. We also modify the image peripheral to produce lower-level representations and divide the encoded images into patches before interacting with other caches. High-level representations used in Omninet might lose spatial signals which can be very useful to help encode other caches. %For example, in order to answer a question about whether an apple in the image is on or under the desk, preserving more spatial signals benefits the interaction between the question and the image.% cross cache attention, which enable spatial, temporal and structured caches interact with each other in the encoding process. Generally, we encode one

Our main contributions are summarized as follows.

\begin{enumerate}
    \item We extend Omninet to handle structured data effectively and to deal with a various number of structured data sources.
    \item We enhance its encoding process with cross cache attention and incorporate the idea of patches to enable cross cache interactions on lower level image representations. 
    \item The proposed model is evaluated on several multimodal datasets, which cover a wide range of modalities, including images, textual inputs, structured data, and videos. It demonstrates a significant improvement against Omninet on all datasets.
\end{enumerate}

The source code is will be disclosed once the paper is accepted.

In Section~\ref{sec:related}, we discuss related work. The proposed model is described in Section~\ref{sec:model}. The datasets and experimental setup are  described in Section~\ref{sec:datasets}. Section~\ref{sec:results} discusses the computational results and the conclusions are drawn in Section~\ref{sec:conclusion}.

\section{Related Work}\label{sec:related}

% \begin{enumerate}
%   \item Multimodal muti-task 
%   \begin{enumerate}
%      \item multimodal learning models tailored to specific tasks/modalities: UniT, MulT, UNITER, VideoBERT, etc.
%      \item More recent models: perceiver io, "attention bottlenecks for multimodal fusion"
%   \end{enumerate}
   
%   \item Fusion
%   \begin{enumerate}
%      \item late fusion
%      \item structured fusion
%      \item cross modality attention (no structured, vision on whole frame level / not on patches)
%   \end{enumerate}
  
%   \item ViT variants (put in background?)
%   \item How our model similar/different from the others
% \end{enumerate}

% 1. Multimodal muti-task 
%     1.1 multimodal learning models tailored to specific tasks/modalities: UniT, MulT, UNITER, VideoBERT, etc.
%     1.2 More recent models: perceiver io, "attention bottlenecks for multimodal fusion"
% 2. Fusion
%     2.1 late fusion
%     2.2 structured fusion
%     2.3 cross modality attention (no structured, vision on whole frame level / not on patches)
% 3. ViT variants (put in background?)
% 4. How our model similar/different from the others

Most existing works in multimodal learning focus on specific tasks with a fixed set of modalities, such as images and text. Many works concatenate image and text inputs and encode them together with Transformer \cite{chen2020uniter,li2020unicoder}. VideoBERT \cite{sun2019videobert} and VisualBERT \cite{li2019visualbert} extend such a transformer-based model to video data. ViLBERT \cite{lu2019vilbert} keeps a separate stream for each modality and enables cross-modality connections to encode one modality with the other. 

Another category of works extends a multimodal learning model in the context of multiple tasks. With the encoder-decoder architecture, MultiModel \cite{kaiser2017one} and UniT \cite{hu2021unit} are able to handle multiple tasks, such as classification and sequence prediction, with just one model. For a single task, MultiModel does not have support for inputs having more than one modality. UniT addresses this issue by encoding images and text with transformers and concatenating encodings from both modalities. It is still limited to only image and language modalities. Perceiver IO \cite{jaegle2021perceiver} is able to handle tasks with different modalities. However, it does not support multiple tasks of various modalities at the same time. To the best of our knowledge, Omninet \cite{pramanik2019omninet} is the most general architecture for multimodal and multi-task learning. Multiple tasks with inputs in different modalities can be trained together in one model. However, it lacks support for structured data. In addition, the lack of interactions between caches limits its ability in the encoding process.  %("UniT: Multimodal Multitask Learning with a Unified Transformer". Concatenate image encodings and text encodings, then feed them to decoder. If one modality is missing, feed encodings of one modality to decoder only. An analogy to Omninet, adding encodings of image and text to temporal cache. Since UniT lacks of breakdowns by spatial and temporal aspect and uses explicit modalities instead, it is limited to only image and language tasks. Omninet can handle other modalities including videos and time series.) However, these models are still limited to a fixed set of modalities.

% The design of spatial and temporal cache makes Omninet capable of handling various tasks and modalities that involving spatial and temporal dimensions. However, no structured and no interactions between caches.

To accommodate structured data, Omninet can be easily extended using late fusion, which is widely used in combining structured data with other features \cite{liu2020fusing,zhang2020combining}. For example, we can encode structured data in one-hot encoding with a few linear layers and concatenate the output with the unstructured feature vector. However, it has a few limitations. The dimension of concatenated structured data depends on the number of structured data sources, which means the model's dimension needs to be changed to adapt to new use cases when the number of structured data sources are different. In addition, the information in the structured data may help in encoding other modalities but late fusion mechanisms do not provide such interactions. Entity embeddings \cite{guo2016entity} have been used to encode structured data \cite{zhu2020location,kulkarni2021pvg}. The encoded structured features are usually concatenated into one feature vector and then combined with other features using late fusion techniques \cite{kulkarni2021pvg}.

Our proposed cross-cache attention mechanism falls into the category of attention-based fusion \cite{zhang2020multimodal}. Previous works using attention-based fusion in multimodal learning focus on vision-language interactions \cite{tsai2019multimodal}. This kind of attention-based fusion has not yet been studied on structured data and image patches in the multimodal multitask learning scenario. 

%Late fusion includes sum, concatenate, ... (see research report). Problem. Structured fusion \cite{zhang2020combining,kulkarni2021pvg}. Early fusion. Cross modality attention. Use entity embedding in late fusion \cite{kulkarni2021pvg,li2019learning} %The weighted sums require dedicated scalar weights. 
%  (references). 

\begin{figure*}
    \centering
    % \captionsetup{justification=centering}

    \includegraphics[width=0.7\textwidth]{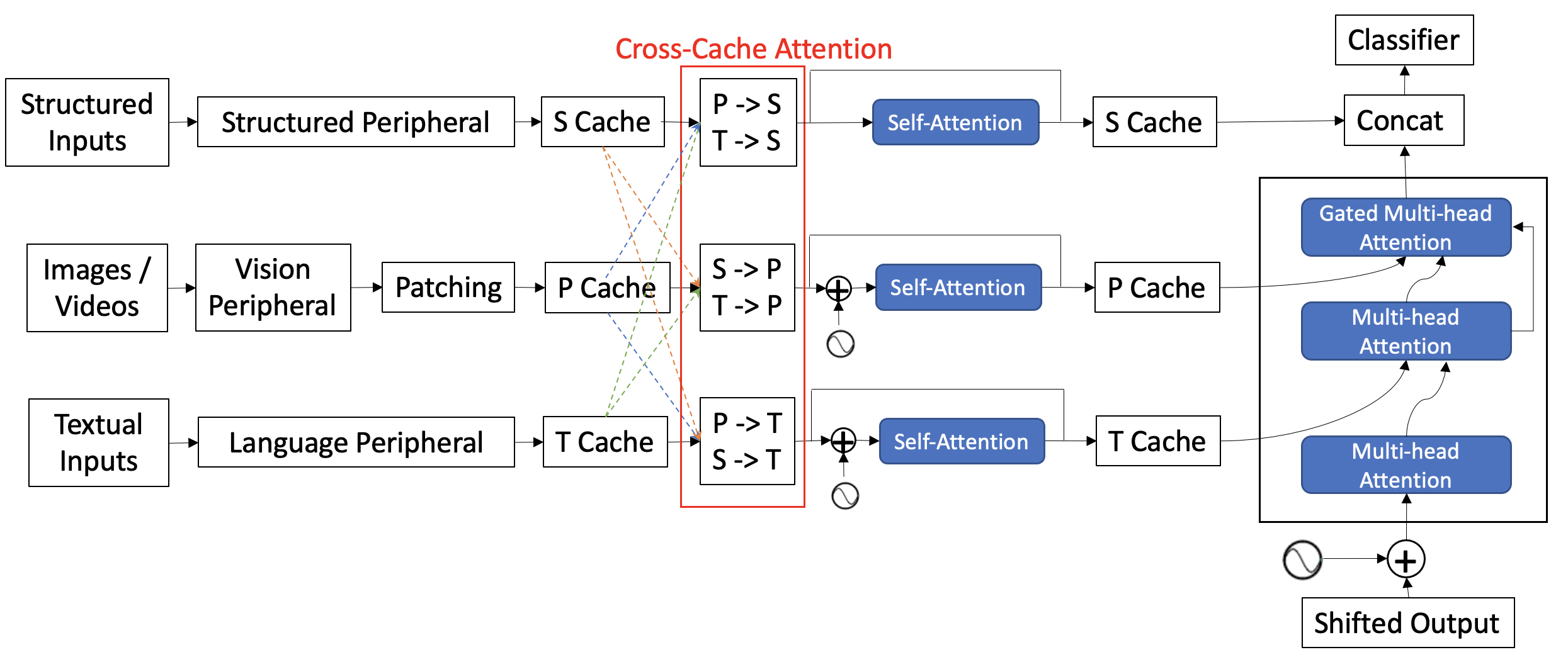}

    \caption{S-Omninet architecture. Caches are denoted as structured (S) cache, spatial (P) cache and temporal (T) cache.}
    \label{fig:cca}
\end{figure*}

\section{Model}\label{sec:model}

Figure~\ref{fig:cca} shows the architecture of the proposed model. A single sample input can be any combination of zero or more of the following modalities: an image $X_{image} \in \mathbb{R}^{H \times W \times C}$, video frames $X_{video} \in \mathbb{R}^{F \times H \times W \times C}$, textual input (i.e., sentences of $Q$ words) and structured data $X_{structured} \in \mathbb{R}^{M}$. For example, a sample can be one or more images, one or more sentences and a video. The number of video frames $F$, length of textual input $Q$ and the number of structured features $M$ can be different across samples.

Omninet has a peripheral for each different modality. A peripheral produces intermediate encodings of the corresponding modality. There are 2 lists. Modalities of spatial matrix are added to the spatial and those corresponding to sequences to the temporal cache list. A sequence of spatial matrices (e.g., video frames) has both spatial and temporal aspects. Omninet encodes each spatial matrix through the spatial peripheral and also stores encodings in the spatial cache. Each matrix encoding is also aggregated through a pooling layer and stored in the temporal cache. A self-attention based temporal encoder is used to calculate embeddings of the temporal cache. Both caches are fed to the Decoder, which is based on the decoder architecture from Transformer \cite{vaswani2017attention} with an additional gated-attention layer to handle the spatial cache. There is also a domain/task encoding. The domain encoding is appended to intermediate encodings from a peripheral to distinguish among different modalities. The task encoding is used as the input of the Decoder to identify different tasks.

% We inherit the following modules from Omninet. Vision peripheral is a pre-trained ResNet-152 model \cite{he2016deep}. Language peripheral uses byte-pair encoding \cite{sennrich2015neural} followed by an embedding layer to generate word embeddings. . For input of multiple modalities, Omninet encodes each modality and stores its encodings in corresponding caches. 

There are 3 aspects that Omninet does not capture: (1) structured data are neither spatial nor temporal and thus cannot be directly modeled; (2) in the embedding phase the two caches are treated independently of each other while often there is interaction, and (3) the spatial cache is used at the pixel level with no notion of locality.

In the following subsections, we introduce in detail the encoder of S-Omninet, including the new structured stream (structured peripheral and structured cache), the enhancement of spatial cache through patching, the cross-cache attention modules and additional self-attention modules on caches other than temporal cache. 

\subsection{Structured Peripheral and Cache} \label{encode_structured}

We propose to use a structured peripheral to encode structured data and put them into the structured cache. In order for the structured cache to effectively communicate with the other two caches, we use entity embeddings to encode categorical features in the structured peripheral. Let us assume there are $C$ categorical features, denoted as $s_1, s_2, ..., s_C$. Each state of a categorical feature is mapped to a vector as $s_i \rightarrow \mathbf{s}_i \in \mathbb{R}^{D}$ through a trainable embedding layer of dimension $D$. It functions as if we ``tokenize'' all possible states of each feature. For example, a color feature can have different embedding vectors for the value `red' and `blue.' Our model may learn similar embeddings for structured value `red' and textual input `red.' We argue that this helps the model match similar concepts between structured and unstructured data.

As the entity embeddings encode each category separately, we may lose useful patterns that can be learned from all structured features as a whole. Therefore, we also keep the encoding of the whole structured data including continuous features besides entity embeddings. The structured peripheral transforms the whole structured sample $X_{structured}$ using one-hot encodings and encodes them through linear layers. The encoded features are denoted as $\mathbf{s} \in \mathbb{R}^{D}$. We then append structured domain encoding to entity embedding features and the traditional whole structured feature and project them back to the dimension $D$ before inserting them into the structured cache. Multiple structured data sources can be encoded and handled by the structured cache. We denote the structured cache containing $N_s$ encodings as  $X_{s} \in \mathbb{R}^{N_s \times D}$.

The structured cache is further encoded in the cross-cache attention modules and self-attention layers, which we introduce later. We use the first encoding of the output from self-attention layers as a representation of the whole structured data and concatenate it with the decoder output before the final prediction.

% The structured cache is further encoded in the cross-cache attention modules, which we will introduce later, and self-attention layers. We use the first element of the cache as a representation of the whole structured data and concatenate it with the decoder output before final the prediction. %Although we use concatenation operation, the model is still capable of handling various numbers of structured data due to the self-attention layers.

\begin{figure*}
    \centering
    % \captionsetup{justification=centering}
    
    \subcaptionbox{Attentions with early self-attention\label{fig:two_ques_early_sa_pt}}{\includegraphics[width=.35\textwidth]{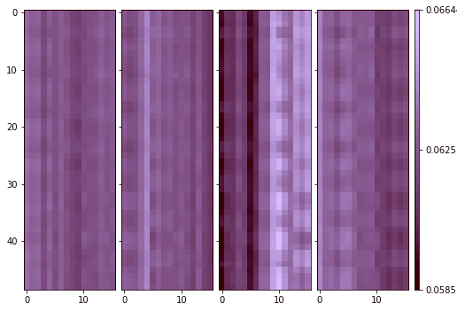}}
    \subcaptionbox{VQA image\label{fig:example_image}}{\includegraphics[width=0.245\textwidth]{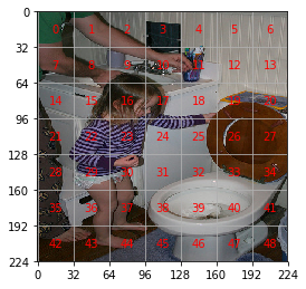}}
    \subcaptionbox{Attentions with late self-attention \label{fig:two_ques_late_sa_pt}}{\includegraphics[width=0.35\textwidth]{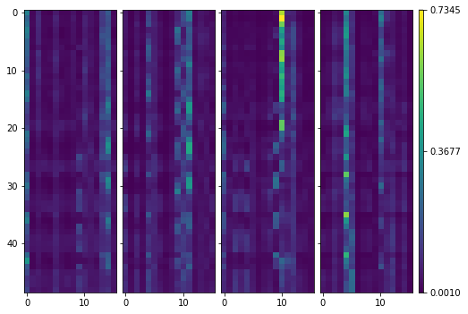}}

    % \begin{subfigure}{0.245\textwidth}
    %     \includegraphics[width=0.2\textwidth]{figs/example_image.png}
    %     \caption{}
    %     \label{fig:example_image}
    % \end{subfigure}
    
    % \begin{subfigure}{0.35\textwidth}
    %     \includegraphics[width=0.2\textwidth]{figs/two_ques_early_sa_pt.png}
    %     \caption{}
    %     \label{fig:two_ques_early_sa_pt}
    % \end{subfigure}
    
    % \begin{subfigure}{0.35\textwidth}
    %     \includegraphics[width=0.2\textwidth]{figs/two_ques_late_sa_pt.png}
    %     \caption{}
    %     \label{fig:two_ques_late_sa_pt}
    % \end{subfigure}
    \caption{An example of VQA image (b) and the attention maps (a, c) in the cross-cache attention $CCA(X_{p}, X_{t})$. In (a) and (c), we plot 4 attention maps, one for each head in the attention. The spatial cache is on the x-axis and the temporal cache is on the y-axis. The question for this image is ``Is the girl potty-trained?'' The encoding of ``girl'' is indexed at 9. The word ``potty'' is encoded with two subwords, ``pot'' and ``ty,'' which are indexed at 10 and 11.}
    \label{fig:two_ques_attn}
\end{figure*}

\subsection{Spatial Cache}

\subsubsection{Patches}

Spatial components of a sample are encoded by a vision peripheral, which produces an $H' \times W' \times d_m$ feature map for an image and $F$ such feature maps for a video, for example. Instead of directly flattening the feature maps and storing them in the spatial cache as done by Omninet, we divide the feature maps into a sequence of 2D patches, similar to ViT \cite{dosovitskiy2020image}. Compared with Omninet, the patches in our model preserve more spatial information than the highly encoded feature maps. 

A patch of a matrix feature map at location $(i,j)$ with height $p_{h}$ and width $p_{w}$ is the rectangle area of the matrix with two diagonal coordinates $(i,j)$ and $(i+p_{h},j+p_{w})$. We obtain patches at each valid location of a matrix feature map with a certain stride. A location is valid if we can obtain a patch without stepping out the boundary. We denote a sequence of patches of a matrix feature map as $\mathbf{z}_{p} \in \mathbb{R}^{T_p \times d_p}$, where $T_p = (H'W')/(p_{h}p_{w})$ is the number of patches and $d_p = (p_{h} \cdot p_{w}) \cdot d_m$. Each patch is mapped to the desired dimension $D$ that matches the size of latent vectors in our attention layers in the spatial cache. All encoded patches of a matrix feature map are denoted as $\mathbf{x}^{0}_{p} = [\mathbf{z}^{1}_{p} \mathbf{E},\mathbf{z}^2_{p} \mathbf{E}, ..., \mathbf{z}^{T_p}_{p} \mathbf{E}]$, where $\mathbf{E} \in \mathbb{R}^{d_p \times D}$. For sequences of matrices (videos), we divide each matrix's feature map into patches and store patches of all matrices in the spatial cache. %$T_p$ is a function of stride $S$. (TODO: write the function).  , tagged with an "IMAGE" domain discriminator and put into the spatial cache. Different from \cite{pramanik2019omninet}, we augment the spatial cache with the awareness of spatial aspect by tokenizing the 2D encoded image $\mathbf{x}_{m} \in \mathbb{R}^{(H \cdot W) \times d_m}$ into a sequence of 2D patches

\subsubsection{Positional Embeddings}
A position embedding is added to each patch to retain position information. A sequence of encoded patches is denoted as $\mathbf{x}_{p} = \mathbf{x}^{0}_{p} + \mathbf{E}_{pos}$, where $\mathbf{E}_{pos} \in \mathbb{R}^{T_p \times D}$. We use learnable positional embeddings as it shows a better performance than the fixed positional embeddings in ViT \cite{dosovitskiy2020image}.  

In the model, the patches may also come from videos, which means temporal relations may exist among patches. In order to capture both the spatial and temporal aspects, two sets of embeddings are learned, each for one of the aspects and each with size $T_p \times \frac{D}{2}$. The spatial embedding $\mathbf{E}^{p}_{pos} \in \mathbb{R}^{T_p \times \frac{D}{2}}$ retains a patch's position information within an image or a video frame. We encode 2D positional information in the spatial embedding and two sets of embeddings are learned, one for each axis. Specifically, we learn X-embedding $\mathbf{E}^{p}_{posx} \in \mathbb{R}^{T_p \times \frac{D}{4}}$ and Y-embedding $\mathbf{E}^{p}_{posy} \in \mathbb{R}^{T_p \times \frac{D}{4}}$. Based on the patch's coordinates, we concatenate the X-embedding and Y-embedding to obtain its spatial embedding. The temporal embedding $\mathbf{E}^{f}_{pos} \in \mathbb{R}^{F \times \frac{D}{2}}$ captures the position of the frame. Then, based on a patch's index of the frame and its position inside the frame we concatenate $\mathbf{E}^{f}_{pos}$ and $\mathbf{E}^{p}_{pos}$ to get the final temporal-spatial embedding $\mathbf{E}_{pos}$. 

Patches are further encoded with the corresponding domain encoding before added to the spatial cache.  We denote the spatial cache with a total of $N_p$ encoded patches as $X_{p} \in \mathbb{R}^{N_p \times D}$.

%We define $P(\mathbf{x}_{m}; p_{h}, p_{w}, S)$ as a patching function on image $X_{m}$ with patch size $(p_{h}, p_{w})$ and stride size $(s_{h},s_{w})$. It produces a list of flattened 2D patches $\bar{X}_{p} = P(\mathbf{x}_{m}; p_{h}, p_{w}, S) \in \mathbb{R}^{T_p \times d_p}$, where $T_p$ is the number of patches and $d_p = (p_{h} \cdot p_{w}) \cdot d_m$. We project each patch to a desired dimension and add a position embedding to it. A sequence of encoded patches is denoted as $X_{p} = [\bar{X}^{1}_{p} \mathbf{E},\bar{X}^{2}_{p} \mathbf{E}, ..., \bar{X}^{T_p}_{p} \mathbf{E}] + \mathbf{E}_{pos}$, where $\mathbf{E} \in \mathbb{R}^{d_p \times d_m}$ and $\mathbf{E}_{pos} \in \mathbb{R}^{T_p \times d_m}$. For video samples, we tokenize each frame into patches and add all encoded patches to the spatial cache. We denote the spatial cache with $N_p$ encoded patches as $X_{p} = [X^{1}_{p}, X^{2}_{p}, ..., X^{T_p}_{p}]$

%, denoted as $p(i,j,p_{h},p_{w})$,
% \begin{displaymath}
%     P(X_{m}; p_{h}, p_{w}) = (p(i,j,p_{h},p_{w})), \forall i \leq H-p_{h}, \forall j \leq W-p_{w}
% \end{displaymath}

\subsection{Temporal Cache}

The temporal cache consists of encodings from domains that have a temporal dimension, such as textual inputs and videos. Textual inputs are encoded first by the corresponding language peripheral and domain encoding. Then they are further encoded by a temporal encoder before put in the temporal cache. For videos, the temporal cache stores frame-level features, which are obtained through pooling from patches. We denote the temporal cache with $N_{t}$ encodings as $X_{t} \in \mathbb{R}^{N_t \times D}$.

% The temporal cache consists of encoded tokens from temporal domains, such as the language and video domain. Each token is associated with a temporal domain discriminator. For video samples, we obtain a encoded token for each frame from a self-attention layer on its patches and add frame tokens to the temporal cache. We denote the temporal cache with $N_{t}$ tokens as $X_{t} = [\mathbf{x}^{1}_{t}, \mathbf{x}^{2}_{t}, ..., \mathbf{x}^{N_{t}}_{t}] \in \mathbb{R}^{N_t \times D}$.

%Encoded text samples vary in length but share the same dimension $d_t$. %Each image or each frame of a video contributes one feature vector to the temporal cache with the "IMAGE" domain discriminator. We denote $N_{t}$ text components of the temporal cache as $C_{t} = (X^{(1)}_{t}, ..., X^{(N_{t})}_{t})$ and $N_{t'}$ vision components as $C_{t'} = (X^{(1)}_{t'}, ..., X^{(N_{t'})}_{t'})$.

\subsection{Cross-cache Attention}

We define the cross-cache attention of cache $X_{\alpha} \in \mathbb{R}^{T_{\alpha} \times d_{\alpha}}$ and $X_{\beta} \in \mathbb{R}^{T_{\beta} \times d_{\beta}}$ as $CCA (X_{\alpha}, X_{\beta}) \in \mathbb{R}^{T_{\alpha} \times D}$. The destination cache $X_{\alpha}$ provides queries and the source cache $X_{\beta}$ provides keys and values. Our model consists of 3 streams of cross-cache attention:

%: $CA_{\beta \rightarrow \alpha} (X_{\alpha}, X_{\beta}) = \text{softmax} (\frac{Q_{\alpha} K_{\beta}^T}{\sqrt{d_{k}}}) V_{\beta}$. We define Querys as $Q_{\alpha} = X_{\alpha} W_{Q_{\alpha}}$, Keys as $K_{\beta} = X_{\beta} W_{K_{\beta}}$ and Values as $V_{\beta} = X_{\beta} W_{V_{\beta}}$, where $W_{Q_{\alpha}} \in \mathbb{R}^{d_{\alpha} \times d_{k}}$, $W_{K_{\beta}} \in \mathbb{R}^{d_{\beta} \times d_{k}}$ and $W_{V_{\beta}} \in \mathbb{R}^{d_{\beta} \times d_{v}}$. We define a cross-cache attention, same as \cite{tsai2019multimodal}, from $\beta$ to $\alpha$ as $CA_{\beta \rightarrow \alpha} (X_{\alpha}. X_{\beta}) \in \mathbb{R}^{T_{\alpha} \times d_v}$: $CA_{\beta \rightarrow \alpha} (X_{\alpha}, X_{\beta}) = \text{softmax} (\frac{Q_{\alpha} K_{\beta}^T}{\sqrt{d_{k}}}) V_{\beta}$.

% \begin{displaymath}
% CA_{\beta \rightarrow \alpha} (X_{\alpha}, X_{\beta}) = \text{softmax} (\frac{Q_{\alpha} K_{\beta}^T}{\sqrt{d_{k}}}) V_{\beta} .
% \end{displaymath}

\begin{displaymath}
\begin{split}
    Y_s &:= concat (CCA(X_{s}, X_{p}), CCA(X_{s}, X_{t})), \\ %\in \mathbb{R}^{N_{s} \times 2D} \\
    Y_t &:= concat (CCA(X_{t}, X_{p}), CCA(X_{t}, X_{s})), \\ %\in \mathbb{R}^{N_{t} \times 2D} \\
    Y_p &:= concat (CCA(X_{p}, X_{s}), CCA(X_{p}, X_{t})) .%\in \mathbb{R}^{N_{p} \times 2D} .
\end{split}
\end{displaymath}

\noindent Different from other cross-modality attention models \cite{tsai2019multimodal,tan2019lxmert}, our model consists of a stream for the structured modality.  Unstructured inputs are broken down into spatial and temporal caches instead of encodings of each modality separately. Additionally, instead of performing cross-modality attention on the pixel-level embeddings, our model captures spatial information during cross-cache attention by taking advantage of the patch features.

% \subsection{Decoder}

% In the current model, we do not modify Omninet's Decoder. After the cross-cache attention, the spatial and temporal cache are consumed by the Decoder in the same way as Omninet. We apply self-attention layers on structured cache and concatenate the last output with the output of Decoder. 

% \begin{figure}[htbp]
%     \centering
%     \captionsetup{justification=centering}

%     \begin{subfigure}{0.5\columnwidth}
%         \includegraphics[width=\textwidth]{figs/parameter_divergence_clients_sp.png}
%         \caption{}
%         \label{fig:dl_sp}
%     \end{subfigure}\hfill
%     \begin{subfigure}{0.47\columnwidth}
%         \includegraphics[width=\textwidth]{figs/parameter_divergence_clients_cov.png}
%         \caption{}
%         \label{fig:dl_cov}
%     \end{subfigure}

%     \caption{Weight divergence among local updates}\label{fig:divergence}
% \end{figure}

\subsection{Late Self Attention}

As shown in Figure~\ref{fig:cca}, we add self-attention layers after cross-cache attentions. In the original design of Omninet, modalities with the temporal dimension are encoded with self-attention based temporal encoder before being put into caches. We initially applied cross-cache attentions after the self-attention layers. However, we found that cross-cache attentions cannot capture interactions effectively. As shown in Figure~\ref{fig:two_ques_early_sa_pt}, we see that all spatial cache encodings have almost the same attention pattern on temporal cache encodings. The attention scores are all very close, ranging from $0.058$ to $0.067$. The attention scores on each row sum up to $1$. For each spatial cache encoding, its attention scores on all $16$ temporal cache encodings are almost the same, close to the average $1/16$ or $0.0625$. The reason is that self-attention makes the encodings in temporal cache similar, i.e., having large cosine similarity. By moving the self-attention layers after the cross-cache attentions, we observe a different pattern, as shown in Figure~\ref{fig:two_ques_late_sa_pt}. It shows that cross-cache attentions pay various attentions to different spatial-temporal encoding pairs. The most relevant pair gets an attention score around $0.7$, while the scores of irrelevant pairs stay well below $0.3$. One encoding getting an attention score of $0.7$ means the other $15$ encodings together get only $0.3$. The attention scores vary much larger than the previous case, see more discussions in Section~\ref{sec:results-vqa}. Note that, MulT \cite{tsai2019multimodal} also put self-attention layers after their cross-modality attentions. The authors empirically argue that this design benefits cross-modality attention without further explanations. %Although we end up with a similar design, the reason behind it may be totally different due to the difference in the inputs our cross-cache and their cross-modality attentions. 

In Omninet, the whole frame encodings are also stored in the temporal cache, so it can learn temporal correlations between frames. We preserve this design in our model. However, this prevents cross-cache attention from learning cross-cache correlations effectively. The reason is that the patch encodings are much closer, in terms of cosine similarity, to the frame encodings than other temporal cache encodings coming from other modalities. This results in high attention scores on the frame encodings, which overshadow the interactions between spatial cache and other temporal cache encodings. Therefore, we exclude the frame encodings in the cross-cache attention.

\subsection{Residual Connections}

Residual connections are commonly used in transformers \cite{vaswani2017attention,dosovitskiy2020image}. We also add residual connections after self-attention blocks to mitigate the vanishing gradient problem as our network is even deeper than Omninet. Additionally, since we put self-attention blocks after cross-cache attentions, the model may lose some cross-cache signals learned previously, which are critical to the predictions in many cases. The residual connections keep the cross-cache signals and merge them with the outputs of self-attention blocks. We observe a significant improvement by adding residual connections. We also verify that the residual connections are active in inference by comparing the weights between the residual connections and the outputs of the self-attention blocks.

%The quality of cross modality attention depends on the variety between encoding vectors. Data of different modalities are usually encoded by different peripherals, e.g., transformer-based language peripheral for text inputs and CNN-based image peripheral for vision inputs. Encoding vectors from the same modality are usually more similar to each other than those from a different modality. As a result, it is hard for the cross modality attention to properly learns similarity across modalities as it is dominated by the similarity between elements of the same modality. For example, we have an image has an apple and a sentence mentioning an apple and a car. We expect the cross modality attention from image to text can help a model to locate the word apple given the elements of the image. However, if the encoding of word apple and car are too similar to each other, it is hard for the modal to correctly match the word apple with the image encoding of apple. %the cross modality attention from Y to X may produce almost the same attention patterns for each element of X.

%The commonly used self attention mechanism for language inputs brings encodings closer, thus making it harder for cross modality attention to learn similarity across modalities. Therefore, we move the self attention layers on language inputs after the cross modality attention.

\subsection{Universal Architecture and Configurations}

S-Omninet is a universal architecture, i.e., we have one and only one model for all tasks. The configuration, such as the sizes of dense layers and the number of attention layers, is also the same across tasks. The dimension of embeddings $D$ in both encoder and decoder is 512. The vision peripheral produces $14 \times 14$ feature maps and the patch size is $2 \times 2$. The self-attention blocks on the temporal cache have 6 layers and 8 heads, the same as Omninet. In other attention blocks, including self-attention blocks on other caches and all cross-cache attention blocks, we use 3 layers and 4 heads. 

Vision and language peripherals are pre-trained and fixed during training of S-Omninet. We train the structured peripheral along with the main architecture of the model. The reason is that different structured data can have very different patterns and it does not make much sense to use a structured peripheral that is pre-trained on totally different features. In addition, the structured peripheral is light and relatively easy to train along with the main model.

For classification tasks, we add dense layers after the decoder as the prediction head. For frame generation tasks, we use the generator model in DCGAN \cite{radford2015unsupervised} and modify the configurations to fit our feature and frame dimensions. Excluding the pre-trained peripherals and prediction heads, S-Omninet and Omninet has 125.4 million and 96.6 million trainable parameters, respectively.

% \begin{figure}
%     \centering
%     \captionsetup{justification=centering}

%     \includegraphics[width=0.6\columnwidth]{figs/example_image.png}

%     \caption{VQA example image}
%     \label{fig:vqa_example}
% \end{figure}

\begin{table}
  \caption{Test performance comparison}
  \centering
  \label{tab:results}
  \begin{tabular}{@{}lS[table-format=2.3]S[table-format=2.2]c@{}}
    \toprule
    Datasets&{Omninet}&{S-Omninet}&Improvement(\%)\\
    \midrule
    VQA & 56.3 & 57.3 & 1.83\\
    S-VQA & 61.1 & 61.7 & 1.01 \\
    Social-IQ & 64.7 & 66.9 & 3.31 \\
    MOSI-Sen & 75.5 & 78.6 & 4.22 \\
    MOSI-Gen & 0.116 & 0.013 & 2.73 \\
    % \multirow{2}{*}{CMU-MOSI} & \multicolumn{1}{c}{XXX} & \multicolumn{1}{c}{XXX} \\%\cline{2-3}
    %                              & \multicolumn{1}{c}{XXX} & \multicolumn{1}{c}{XXX} \\%\cline{2-3}
  \bottomrule
\end{tabular}
\end{table}

\section{Datasets}\label{sec:datasets}

\paragraph{Social-IQ} Social-IQ \cite{zadeh2019social} is a video question answering dataset that contains 1,250 annotated videos, 7,500 questions and 52,500 answers. Each question is provided with 4 correct answers and 3 incorrect answers. All answers are sentences. The task is to predict whether an answer is correct given a video with a question. We extract video frames at 1fps \cite{zadeh2019social} and each video sample consists of 55 frames on average. 

\paragraph{CMU-MOSI} CMU-MOSI \cite{zadeh2016multimodal} is a multimodal human sentiment dataset. It consists of 2,199 video clips of faces during conversations. Each video clip is labeled with a sentiment score between -3 and 3. We evaluate the model performance on two tasks. One is MOSI-Sen, a binary classification task to predict whether the sentiment is positive or negative using video frames and transcripts. The transcripts, text translated from video's audio, are provided in the dataset. The other task is MOSI-Gen where we generate the next frame of a video clip, given previous frames and the transcript of this clip. We use accuracy for the classification task and Mean Absolute Error (MAE) for the frame generation task.

\begin{figure}
    \centering
    \captionsetup{justification=centering}

    \includegraphics[width=0.48\columnwidth]{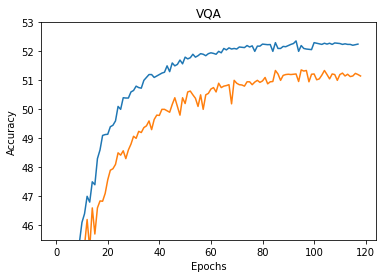}
    \includegraphics[width=0.49\columnwidth]{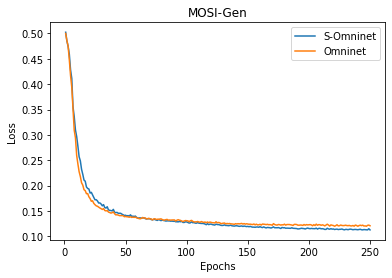}
    
    \caption{Learning curves on the validation set}
    \label{fig:val_compare}
\end{figure}

\paragraph{Visual Question Answering} The VQA v2.0 dataset \cite{goyal2017making} is a large visual question-answering dataset consisting of approximately 1.1 million (image, question) pairs with 13 million answers on MSCOCO images. Every question is associated with two similar images that result in two different answers. We evaluate the model performance on the provided test-dev set same as Omninet \cite{pramanik2019omninet}. 

\paragraph{Structured \& Visual Question Answering} Due to the lack of a public multimodal dataset that contains structured data and multiple unstructured data, we create the S-VQA dataset which contains images, text and structured data. In practical business processes, this type of data is very often encountered which motivates this study. A single model for a given process serves different specific use cases. Unfortunately, the lack of established multi-modal approaches, combined with such data sets being proprietary means no well-curated data sets are available. Therefore, we create the S-VQA that contains a large number of multimodal samples. 

The key part is to create structured data that have a meaningful interaction with other modalities to simulate the real-world cases. The structured data are composed of both numerical and categorical features. We create it in such a way that the numerical features are correlated with the labels and categorical features are correlated with both the labels and the unstructured data. For numerical features, we randomly sample $n_c$ vectors with the desired dimension and treat them as the \textit{centroids}. The value of $n_c$ is the same as the number of classes in the VQA v2.0 dataset. Given a covariance matrix, we sample random vectors around each \textit{centroid} from Gaussian distributions. Vectors sampled from the same \textit{centroid} are assigned with the same class label. A total of 3,500 clusters are generated to be matched with all classes. For categorical features, we first identify important elements (words in a sentence or regions of an image) in existing modalities. The importance of each element is determined by the attention score produced by the decoder of the vanilla Omninet during training. Then, we create categorical features by clustering the important elements. Each cluster is considered a categorical feature. In addition, we perturb the generated structured features to introduce correlations to the labels. More details are in Appendix~\ref{a:svqa}.

\section{Results}\label{sec:results}

We build S-Omninet on top of Omninet by adding implementations of the cross-cache attention module, the patching module and the structured peripheral. For the vision peripheral, we use a pre-trained ResNet-152 model \cite{he2016deep} with the last pooling layer and a few convolution layers removed to get the desired $14 \times 14$ feature maps. We use the same language peripheral as Omninet. We run experiments on NVIDIA GeForce RTX 2080 Ti GPUs.

Table~\ref{tab:results} shows the performance comparison between our model and the baseline. On the VQA and S-VQA datasets, we observe an improvement of 1.83\% and 1.01\% on the accuracy, respectively. On the social-IQ dataset, our model achieves an accuracy of 66.9, which is 3.31\% better than Omninet. The accuracy scores of both our model and Omninet are higher than 63.91, which is a baseline performance as reported in the original paper \cite{zadeh2019social}. On MOSI-Sen and MOSI-Gen, S-Omninet is 4.22\% better in accuracy on the sentiment prediction task and 2.73\% better in MAE on the frame prediction task.

Figure~\ref{fig:val_compare} shows the validation accuracy curves on VQA and validation loss curves on MOSI-Gen. S-Omninet has a similar converging behavior as Omninet. The curves on the other datasets demonstrate a similar pattern. However, since our model is slightly larger than Omninet, the actual training time is 23\% longer per epoch than Omninet.

\begin{figure}
    \centering
    % \captionsetup{justification=centering}
    
    \subcaptionbox{}{\includegraphics[width=0.45\columnwidth]{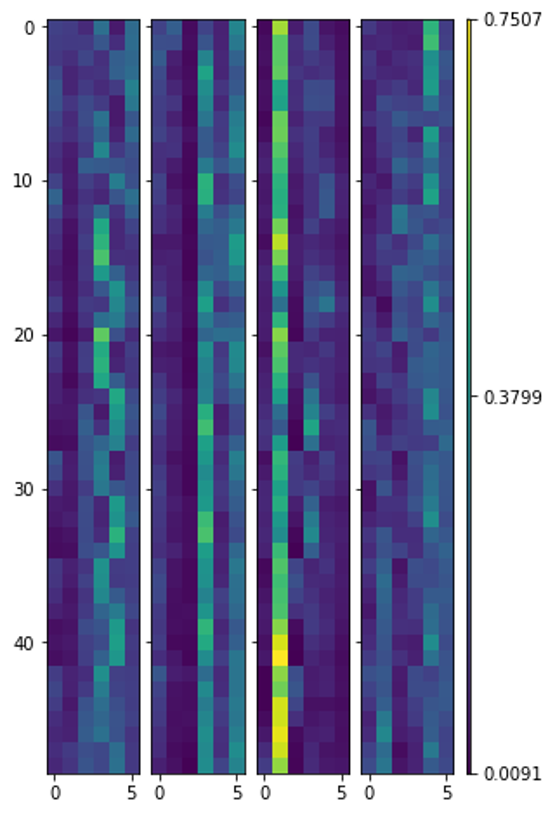}}
    \subcaptionbox{}{\includegraphics[width=0.225\columnwidth]{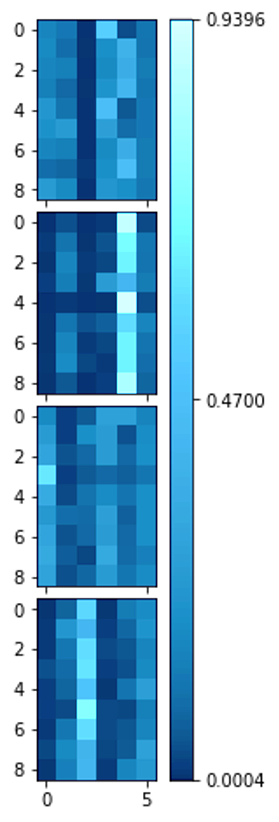}}

    \caption{Cross-cache attention on structured features. The structured cache is on the x-axis and the other cache is on the y-axis. Left: attention maps in $CCA(X_{p}, X_{s})$. Right: attention maps in $CCA(X_{t}, X_{s})$.}
    \label{fig:svqa_attn}
\end{figure}

\subsection{VQA}\label{sec:results-vqa}

As shown in Table~\ref{tab:results}, our model shows significant improvements over Omninet. Omninet's performance on the VQA dataset was first reported as 55.3 \cite{pramanik2019omninet}. With more tuning, we achieve an accuracy of 56.3 and use this model as the baseline. Our model improves the baseline by 1.83\%, which demonstrates the benefits of adding cross-cache attention. 

We also test our model in a more challenging task, ``two-question VQA,'' which demonstrates the effectiveness of cross-cache attention more clearly. In this task, we randomly add an irrelevant question before or after the original question. For example, the textual input of the example in Figure~\ref{fig:two_ques_attn} is ``What is this a collection of? Is the girl potty-trained?'' The second sentence is the original question for this image and the first sentence is an irrelevant question. 

On this ``two-question'' VQA dataset, we argue that Omninet cannot tell which question is the relevant one due to the lack of interactions between caches. Since each cache is encoded separately, Omninet cannot use spatial information to identify the relevant question. Therefore, the performance of Omninet is substantially impacted and the accuracy drops from 51\% to 41\%. In contrast, our model with cross-cache attention shows a good ability to identify the relevant question. Figure~\ref{fig:example_image} shows an example of a VQA input and Figure~\ref{fig:two_ques_late_sa_pt} shows the attention maps in cross-cache attention on this example. In the cross-cache attention $CCA(X_{p}, X_{t})$, image patches are encoded with the textual features. As shown in Figure~\ref{fig:two_ques_late_sa_pt}, words in the irrelevant question, indexed from 0 to 8 on the x-axis, get low attention scores. As a result, our model shows an 8\% improvement in accuracy over Omninet on this challenging task.

Besides paying more attention to the correct sentence, cross-cache attention also stresses more relevance to more relevant words. We can see that in all 4 heads, the words ``girl'' and ``potty'' have a high attention score in many cases. Especially in the second head, the word ``girl'' and ``potty'' are linked with the regions that show the girl and potty. Note that, we mark grids in the original image to ease the visualization, but the grid indices are not strictly mapped to patch indices on the y-axis. The reason is that the patch embeddings used in cross-cache attention come from CNN-encoded feature maps. Each patch also contains information in its surrounding patches due to the convolution operations.

\subsection{Cross-cache Attention on Structured Data}

Omninet does not handle structured data. In order to train Omninet on S-VQA, we encode the structured data with one-hot encodings followed by a linear layer. Then the encoded structured data is concatenated with the output of the decoder. On the S-VQA dataset, Omninet achieves an accuracy of 61.1\%, higher than the accuracy on the VQA dataset, as the structured data provides extra information about the labels. Our model further improves Omninet by 1.01\% relatively. In the structured cache, the first encoding is the embedding of the whole structured feature vector including numerical and categorical features. The second encoding is the embedding of categorical features generated from important spatial input signals. The rest are encodings of categorical features of important temporal inputs. Figure~\ref{fig:svqa_attn} shows the attention maps on cross-cache attention $CCA(X_{p}, X_{s})$ and $CCA(X_{t}, X_{s})$, where the model encodes each cache with more attention on corresponding structured encodings. For example, the spatial cache is encoded with more attention on the second structured encoding and the temporal cache is encoded with more attention on the third and fifth structured encodings. It demonstrates the effectiveness of cross-cache attention in integrating correlated structured and unstructured data.

\begin{figure}
    \centering
    \captionsetup{justification=centering}

    \subcaptionbox{Omninet\label{fig:mosi-gen-omninet}}{\includegraphics[width=0.32\columnwidth]{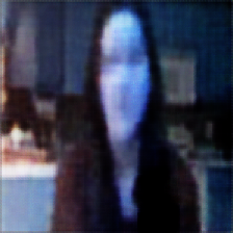}}
    \subcaptionbox{CCA\label{fig:mosi-gen-cca}}{\includegraphics[width=0.32\columnwidth]{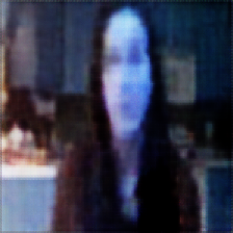}}
    \subcaptionbox{Real\label{fig:mosi-gen-target}}{\includegraphics[width=0.32\columnwidth]{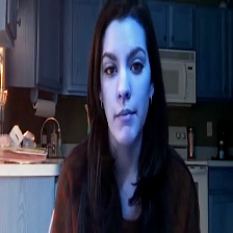}}

    \caption{Generated frames from different models}
    \label{fig:mosi-gen}
\end{figure}

\subsection{Video Tasks}

Social-IQ and CMU-MOSI are both video datasets with different tasks. On classification tasks, we observe significant improvements against Omninet. Our model is 3.3\% and 4.2\% better than Omninet on Social-IQ and MOSI-Sen, respectively. The cross-cache attention modules enable caches to interact with each other, which helps the model locate related encodings in different caches as we see in the ``two-question VQA'' example. This module can be more useful on videos than images because there are multiple frames in a video clip and the model needs to figure out in which frame encodings are more relevant. 

Although Omninet has the Gated Multihead Attention, which makes the model focus more on important frames by increasing the attention scores on them, there are several limits to this mechanism. First, the importance of a frame is calculated from an attention module on the temporal cache, which includes the frame and word embeddings. The frame-level embedding is a highly encoded image feature, where detailed spatial information, which can be critical to deciding whether a frame is important, is missing. Second, once a frame is identified as an important frame, all encodings in this frame get a higher weight in the next attention. Because of that, a less important encoding in an important frame can have a higher weight than an important encoding in a less important frame. The cross-cache attention overcomes these problems by performing attentions in a finer granularity, i.e., directly on the encoding level. 

On the frame generation task, our model shows a 2\% improvement on MAE and we observe that our model generates more eye appealing images, as shown in Figure~\ref{fig:mosi-gen}. The image generated by our model is darker on the eyes, nose and mouth, which shows clearer boundaries of facial features. Both frames that are generated by Omninet and our model are not very crisp. This is often seen on transformer-based models with a simple image generator \cite{jaegle2021perceiver} (this work's focus is not on generating high resolution and sharp images). In spite of that, we still see that the image from our model has additional facial details than Omninet.

\section{Conclusion}\label{sec:conclusion}

In this work, we extend and improve Omninet by introducing cross-cache attention, integrating patch embeddings for vision inputs, and supporting structured data. We discuss the design choice of putting cross-cache attention before self-attentions. In addition, we study the impact of this design and demonstrate reasons it works. The proposed S-Omninet is shown capable of learning structured data of various lengths effectively with unstructured data. It demonstrates the effectiveness of cross-cache attention by showing a significant improvement over Omninet on several multimodal datasets. 

\bibliographystyle{named}
\bibliography{main}

\clearpage

\appendix

\section{Synthetic Structure Data}\label{a:svqa}

\subsection{Categorical Features}

We start from an existing multimodal dataset, e.g., VQA 2.0 \cite{goyal2017making}, and create structured samples with categorical features so that they are correlated with samples of the existing modalities.

We impose correlations between structured and unstructured samples as follows. First, we identify important elements of each existing modality. The importance of each element is determined by the attention score produced by the decoder of the vanilla Omninet during training. For each text input, we identify $P$ most important words. For each image, we find $Q$ most important regions (i.e., low-resolution pixels of the feature map produced by the peripheral). 

Then, we create categorical features given the important elements. For text data, we cluster important words and consider each cluster as a categorical feature. Each word is mapped to a category of a feature. 

Similar to text data, we cluster importance regions and each cluster is considered a categorical feature. For each cluster, we further find sub-clusters and then map each sub-cluster to a category of a feature.

For each VQA sample, we create structured features by assigning a category to each feature according to the important regions/words of the text data and image. If multiple words or regions belong to the same feature, we use the one with a higher importance score. A special value is assigned to represent an empty category. We generate 5 categorical features, one feature is correlated with spatial inputs, and the rest of the 4 features are correlated with the text inputs. We limit the number of categorical features to 5 in order to control the rate of an empty category in a low range (less than 20\%).

\subsection{Perturbing Structured Features}

We perturb structured features to introduce correlations with labels. We consider samples $D$ of class $k$ and feature $X$ with $N$ categories. We have the categorical distribution of $X$: 

\begin{displaymath}
x = (p(X=1), p(X=2), ..., p(X=N)),
\end{displaymath}

\noindent where $p(X=j)=\frac{Count(j)}{|D|}$. First, we generate a Dirichlet prior $Dir(f(k,N))$ depending on classes, where $f(k,N)$ is a function that produces a unique category distribution over $N$ categories for class $k$. For example, if we have 2 classes and a feature with 4 categories, we can have $f(0,4)=(0.7,0.1,0.1,0.1)$ and $f(1,4)=(0.1,0.7,0.1,0.1)$. If we have more classes than categories, we use one or two major categories to differentiate each class. For example, if we have 4 classes and a feature with 3 categories, we can have $f(0,3)=(0.8,0.1,0.1)$, $f(1,3)=(0.1,0.8,0.1)$, $f(2,3)=(0.1,0.1,0.8)$ and $f(3,3)=(0.45,0.45,0.1)$. We create a pool of combinations from $N$ categories, which contains a 1-combination and a 2-combination. For each $(k,N)$ pair, we randomly select one combination of categories that has not yet been selected and mark them as the major categories for this pair. We first select single category combinations and then select combinations of two categories. Then we set $f(k,N)$ based on categories. We set $0.8$ for the major categories and $0.1$ for others, and then normalize them so they sum up to one. Then we draw a sample $y=(y_1,y_2,...,y_{N})$ from $Dir(f(k,N))$ with $\sum_{i=1}^{N} y_i = 1$. Next, we perturb $x$ with $y$ as $z = \frac{x+y}{2}$. Finally, for each sample $i$ in $D$, we draw a category given probability $z$ and assign it to feature $X$.

% \begin{displaymath}
% \begin{split}
%     &x = (x_o, x_s) \\
%     &x_o, x_s \in \mathcal{R}^{512} \\
%     &x' = w^T x + b \\
%     &y = \text{Linear}(x') \\
%     &L = \sum_{i} l(y^{i},x^{i},w) + \lambda \sum_{i=513}^{1024} |w_{i,:}|
% \end{split}
% \end{displaymath}

% \section{\LaTeX{} and Word Style Files}\label{stylefiles}

\end{document}